\documentclass[letterpaper, 10 pt, journal, twoside]{ieeetran}

\usepackage{hyperref}

\usepackage{booktabs,multirow}
\usepackage{algorithm}
\usepackage[noend]{algpseudocode}
\usepackage[english]{babel}
\usepackage[pdftex]{graphicx}
\usepackage{adjustbox}
\DeclareGraphicsExtensions{.pdf,.jpg,.jpeg,.png}

\usepackage{graphicx} 
\usepackage{color}
\usepackage{bm}
\usepackage[caption=false]{subfig}
\usepackage{amsmath} 
\usepackage{amssymb}  
\usepackage{physics}
\usepackage[table]{xcolor}
\usepackage{algcompatible}
\DeclareMathOperator*{\argmin}{arg\,min}
\algblockdefx{FORALLP}{ENDFAP}[1]%
  {\textbf{for all }#1 \textbf{do in parallel}}%
  {\textbf{end for}}

\usepackage{bm}

\usepackage{tikz}

\newcommand\copyrighttext{%
  \footnotesize \textcopyright 2020 IEEE. Personal use of this material is permitted.
  Permission from IEEE must be obtained for all other uses, in any current or future
  media, including reprinting/republishing this material for advertising or promotional
  purposes, creating new collective works, for resale or redistribution to servers or
  lists, or reuse of any copyrighted component of this work in other works.}
  
\newcommand\copyrightnotice{%
\begin{tikzpicture}[remember picture,overlay]
\node[anchor=south,yshift=10pt] at (current page.south) {\fbox{\parbox{\dimexpr\textwidth-\fboxsep-\fboxrule\relax}{\copyrighttext}}};
\end{tikzpicture}%
}

\graphicspath{{./figures/}} 

\title{Memory of Motion for Warm-starting \\ Trajectory Optimization}

\author{Teguh Santoso Lembono$^{1,2}$, Antonio Paolillo$^{1}$, Emmanuel Pignat$^{1,2}$,  and Sylvain Calinon$^{1,2}$
\thanks{$^1$ Idiap Research Institute, Martigny, Switzerland \quad $^2$ EPFL, Lausanne, Switzerland \texttt{\{name.surname\}@idiap.ch}}
\thanks{This work was supported by the European Commission's Horizon 2020 Programme (MEMMO project, http://www.memmo-project.eu/, grant 780684).}
}

\author{Teguh Santoso Lembono$^{1}$, Antonio Paolillo$^{2}$, Emmanuel Pignat$^{1}$, and Sylvain Calinon$^{1}$%
\thanks{Manuscript received: September, 10, 2019; Revised December, 31, 2019;
Accepted February, 03, 2020.}
\thanks{This paper was recommended for publication by Editor Dongheui Lee upon
evaluation of the Associate Editor and Reviewers' comments. This work was supported by the European Commission's Horizon 2020 Programme (MEMMO project, http://www.memmo-project.eu/, grant 780684).}
\thanks{$^{1}$T. S. Lembono, E. Pignat, and S. Calinon are with Idiap Research Institute, Switzerland and EPFL, Switzerland
{(email: teguh.lembono@idiap.ch; emmanuel.pignat@idiap.ch; sylvain.calinon@idiap.ch)}}%
\thanks{$^{2}$A. Paolillo is with Idiap Research Institute, Switzerland
{(email: antonio.paolillo@idiap.ch)}}%
\thanks{Digital Object Identifier (DOI): 10.1109/LRA.2020.2972893}
}

\begin{document}

\maketitle

\copyrightnotice

\begin{abstract}
Trajectory optimization for motion planning requires good initial guesses to obtain good performance. In our proposed approach, we build a memory of motion based on a database of robot paths to provide good initial guesses. The memory of motion relies on function approximators and dimensionality reduction techniques to learn the mapping between the tasks and the robot paths. Three function approximators are compared: $k$-Nearest Neighbor, Gaussian Process Regression, and Bayesian Gaussian Mixture Regression. In addition, we show that the memory can be used as a metric to choose between several possible goals, and using an ensemble method to combine different function approximators results in a significantly improved warm-starting performance. We demonstrate the proposed approach with motion planning examples on the dual-arm robot PR2 and the humanoid robot Atlas. 
\end{abstract}


\section{Introduction}
\label{sec:introduction}

\IEEEPARstart{M}{otion} planning for robots with high Degree-of-Freedoms (DoFs) presents many challenges, especially in the presence of constraints such as obstacle avoidance, joint limits, etc. To handle the high-dimensionality and the various constraints, many works~\cite{schulman2013finding}~\cite{zucker2013chomp}~\cite{kalakrishnan2011stomp} focus on \emph{trajectory optimization} methods that attempt to find a locally optimal solution. In this approach, the motion planning problem is formulated as an optimization problem
\begin{equation}
\min_{\bm{q}_{0:T}} \ell ( \bm{q}_{0:T}), \quad \text{s.t.} \quad \bm{g}( \bm{q}_{0:T}) \leq \bm{0},\quad \bm{h}( \bm{q}_{0:T}) = \bm{0},
\label{eq:minimization}
\end{equation}
where $\bm{q}_{0:T}$ denotes the robot's configurations from time step $t = 0$ to $t = T$; $\ell(\cdot)$,  $\bm{g}(\cdot)$ and $\bm{h}(\cdot)$ are the cost, the inequality and the equality constraints. The solution of~\eqref{eq:minimization} is the \emph{path} $\bm{y} = \bm{q}_{0:T}^{\ast} = ( \bm{q}_{0}^{\ast \top}, \dots, \bm{q}_{T}^{\ast \top})^\top \in \mathbb{R}^{DT}$, with $D$ the dimension of $\bm{q}_{t}$. When the path is parameterized by time, it is called \emph{trajectory}.

\begin{figure}
\centering
\subfloat[][]{\includegraphics[height=0.3\columnwidth]{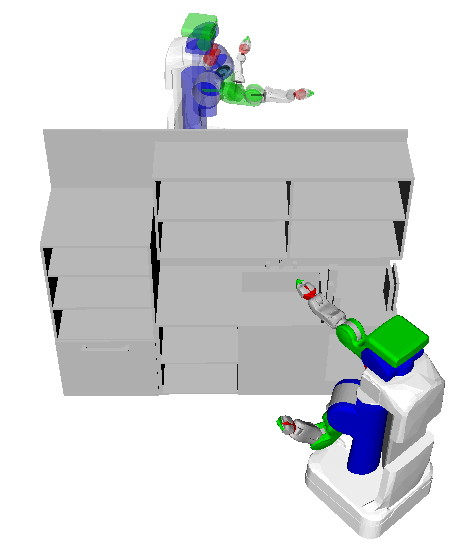}\label{fig:pr2_example_kitchen}}
\hfill
\subfloat[][]{\includegraphics[height=0.3\columnwidth]{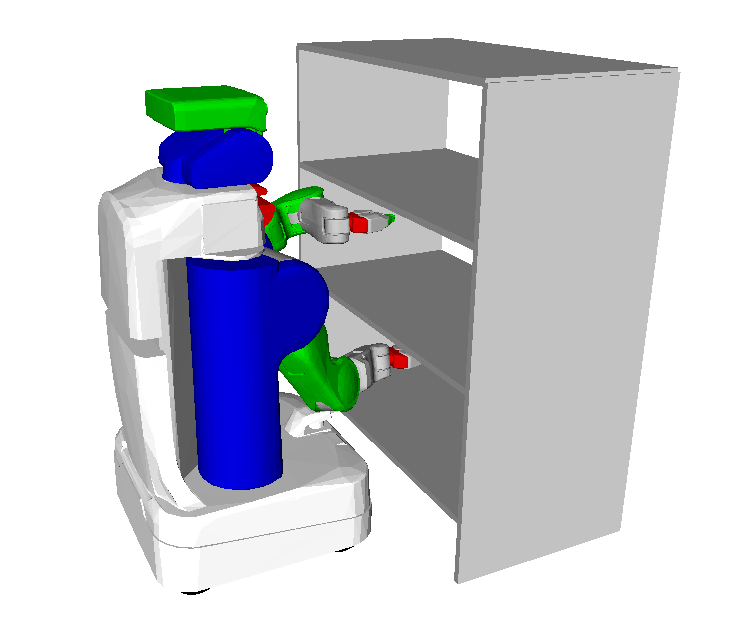}\label{fig:pr2_example_shelf}}
\hfill
\subfloat[][]{\includegraphics[height=0.3\columnwidth]{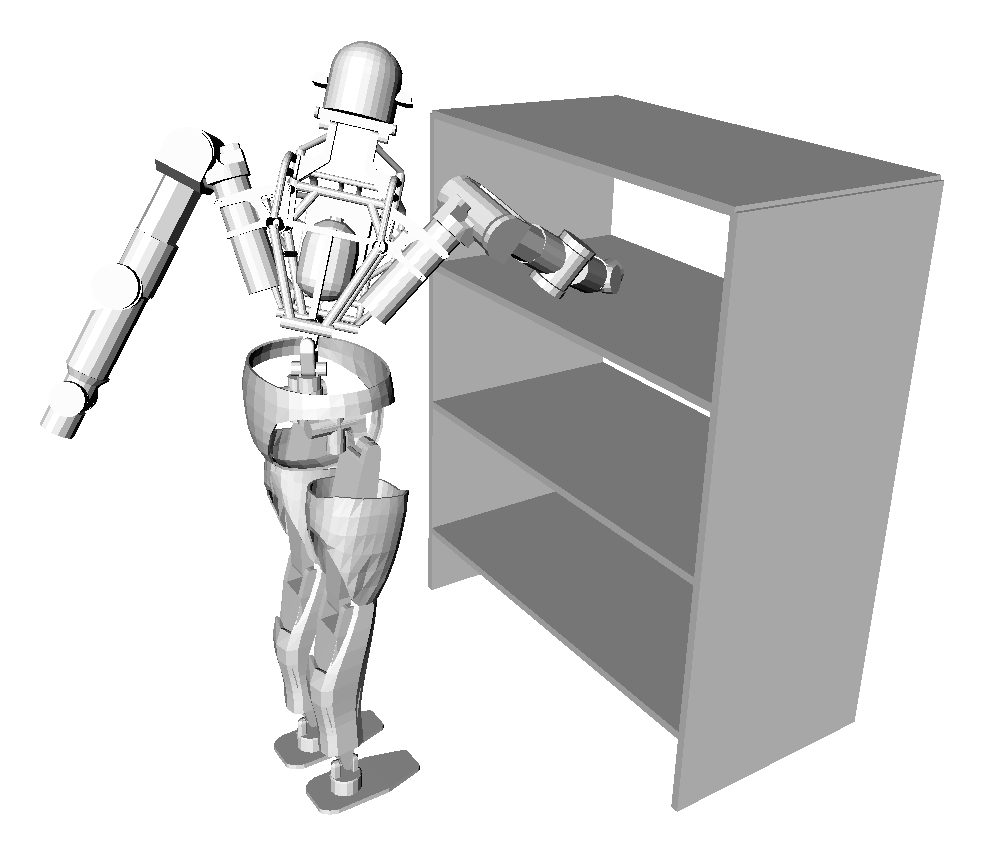}\label{fig:planning_atlas}}
\caption{
Examples of motion planning problems: \protect\subref{fig:pr2_example_kitchen}~moving the PR2 base between two points while avoiding an obstacle, \protect\subref{fig:pr2_example_shelf}~dual arm motion of PR2 to pick items from a shelf to another, and \protect\subref{fig:planning_atlas}~whole body motion of Atlas.}
\label{fig:pr2_example}
\end{figure}

As an example, consider the planning problem depicted in Fig.~\ref{fig:pr2_example}, where the PR2 robot has to move its base around an object or to perform a dual-arm motion to pick items from the shelves. If the task $\bm{x} = {(\bm{q}_\text{init}^\top,\bm{q}_\text{goal}^\top)}^\top$ is to move from an initial configuration $\bm{q}_\text{init}$ to a goal configuration $\bm{q}_\text{goal}$ while minimizing the total joint velocity, the optimization problem can be written as
\begin{equation}
\min_{\bm{q}_{0:T}} \sum_{t=0}^{T-1} \big\|(\bm{q}_{t+1} - \bm{q}_{t})\big\| ^2, \quad \text{s.t.} \quad  \bm{q}_{0} = \bm{q}_\text{init}, \quad \bm{q}_{T} = \bm{q}_\text{goal}.
\label{eq:discrete_vel}
\end{equation}
Other constraints can also be added, e.g.~to avoid collisions, to comply with joint limits, etc. 

Such optimization problems are in general non-convex, especially due to the collision constraints, which makes finding the global optimum very difficult. Trajectory optimization methods such as TrajOpt~\cite{schulman2013finding}, CHOMP~\cite{zucker2013chomp}, or STOMP~\cite{kalakrishnan2011stomp} solve the non-convex problem by iteratively optimizing around the current solution. While such approach is very popular and yields good practical results, the convergence and the quality of the solution are very sensitive to the choice of the initial guess. If it is far from the optimal solution, the method can get stuck at a poor local optimum.

To overcome this problem, our approach builds a \emph{memory of motion} that learns how to provide good initializations (i.e., a \emph{warm-start}) to the solver based on previously solved problems.  Functionally, the memory of motion is expected to learn the mapping $\bm{f}: \bm{x} \rightarrow \bm{y}$ that maps each task $\bm{x}$ to the robot path $\bm{y}$. Such mapping can be highly nonlinear and \emph{multimodal} (i.e., one task $\bm{x}$ can be associated to several robot paths $\bm{y}$), and the dimension of $\bm{y}$ is typically very high. Our proposed method relies on machine learning techniques such as function approximation and dimensionality reduction to learn this mapping effectively. We use the term \emph{memory of motion} to include both the database of motions and the algorithms to query the warm-starts from the database.

We point out that while other techniques such as sampling-based motion planners can also be used to warm-start the solver (e.g. in~\cite{merkt2018leveraging}), such methods typically require a considerable computation time (i.e. in the order of seconds) that is comparable to the solver's convergence time itself, given the very high dimensional problems considered here. In contrast, querying the memory of motion can be done very fast, in the order of milliseconds. Additionally, our proposed method produces initial guesses that are close to the optimal solutions, reducing the convergence time. 

The contribution of this paper is the following. First, we propose the use of function approximation methods to learn the mapping $\bm{f}(\bm{x)}$. We consider three methods: $k$-Nearest Neighbor ($k$-NN), Gaussian Process Regressor (GPR) and Bayesian Gaussian Mixture Regression (BGMR), and discuss their different characteristics on various planning problems. We show in particular that BGMR handles multimodal output very well. Furthermore, we show that the memory of motion can be also be used as a metric for choosing optimally between several possible goals. Finally, we demonstrate that using an ensemble of function approximators to provide warm-starts boosts the success rate significantly.

The paper is organized as follows. In Section \ref{sec:related_works} we discuss the related work that use the concept of memory of motion for various problems. Section \ref{sec:method} explains the methods for constructing and using the memory of motion. The experimental results are presented and discussed in Section \ref{sec:experiments} and \ref{sec:discussion}. Finally, Section \ref{sec:conclusion} concludes the paper.

\section{Related Work}
\label{sec:related_works}

The idea of using a memory of motion to warm-start an optimization solver has previously been explored in the context of optimal control and motion planning. In~\cite{stolle2006policies} a trajectory library is constructed to learn a control policy. A set of trajectories are planned offline using $A^*$ algorithm and stored as library, then $k$-NN is used online to determine the action to perform at each state. In~\cite{liu2009standing}, they use similar approach to predict an initial guess for balancing a two-link robot, which is then optimized by Differential Dynamic Programming. An iterative method to build a memory of motion to initialize an optimal control solver is proposed in~\cite{mansard2018}. They use neural networks to approximate the mapping from the task descriptors (initial and goal states) to the state and control trajectories. Another neural network is trained to approximate the value function, which is then used as a metric to determine how close two states are dynamically. In~\cite{forte2012line} GPR is used to predict new trajectories based on the library of demonstrated movements encoded as Dynamic Movement Primitives (DMP)~\cite{ijspeert2002learning}. GPR is used to map the task descriptors to the DMP parameters.

In robot motion planning, Probabilistic Roadmap (PRM)~\cite{kavraki1998probabilistic} can be seen as the construction of memory of motion by precomputing the graph connecting robot configurations. Some works exploit Rapidly-exploring Random Trees (RRT)~\cite{lavalle1998rapidly}, another popular sampling-based method. For example, in~\cite{martin2007offline} an offline computation is used to speed-up the online planning in the form of an additional bias in sampling the configurations. In ~\cite{phillips2012graphs}, an \emph{Experience Graph} is built from previously planned solutions. During the online planning, the search is biased towards this graph. 
The \emph{Lightning} framework is proposed in~\cite{berenson2012robot} to plan paths in high-dimensional spaces by learning from experience. The path library is constructed incrementally. Given the current path library and a task to be executed, the algorithm runs two versions of the planner online, one that plans from scratch and the other one initialized by the library. In~\cite{jetchev2009trajectory} a high-dimensional (791) task descriptor is constructed, and the metric between the task descriptors is refined to minimize the necessary refinement of the initial trajectory using $L_1$ norm, resulting in a sparse metric and hence sparse descriptors. In~\cite{merkt2018leveraging}, a subindexing is used to reduce the amount of memory storage and to use the subtrajectories of the original solutions. In robot locomotion~\cite{werner2015generalization}, a mapping from the task space to the optimal trajectory for cyclic walking is learned using various machine learning algorithms, but the prediction is not re-optimized online.
In \cite{lampariello2011trajectory}, the initial trajectories for real-time catching are predicted using $k$-NN, Support Vector Regression, and GPR. 

As compared to the above works, our proposed method has the following differences: (i) none of the above methods attempt to handle the multimodal output cases. We show that BGMR can handle well such cases, (ii) we show that the memory of motion can be used as a metric for choosing optimally between several possible goals, and (iii) we show that using an ensemble of methods to provide the warm-start outperforms the individual methods significantly.

\section{Method}
\label{sec:method}

Section \ref{sec:building_memmo} discusses the main idea of building the memory of motion using function approximation and dimensionality reduction techniques to learn the mapping between the task and the associated robot path. Section \ref{sec:metric} explains how the memory of motion can be used as a metric for choosing between different goals. Finally, Section \ref{sec:ensemble} explains how the warm-starting performance can be improved significantly using an ensemble method.

\subsection{Building a Memory of Motion}
\label{sec:building_memmo}

To learn the mapping $\bm{f}: \bm{x} \rightarrow \bm{y}$, we firstly generate a set of tasks $\{\bm{x}\}$ and the corresponding robot paths $\{\bm{y}\}$. This is done by sampling $\bm{x}$ from a uniform distribution covering the space of possible tasks and run the trajectory optimizer to obtain the robot paths $\bm{y}$ until we obtain $N$ samples ($\bm{x}$,$\bm{y}$). Let $\bm{X} = (\bm{x}_0, \dots, \bm{x}_{N-1})$ and $\bm{Y} = (\bm{y}_0, \dots, \bm{y}_{N-1})$. The mapping $\bm{f}$ can be learned by training function approximators using the database $\{\bm{X},\bm{Y}\}$. In this paper we consider three function approximators: $k$-NN, GPR, and BGMR.

\subsubsection{$k$-Nearest Neighbor ($k$-NN)}\label{sec:K-near}

$k$-NN is a very simple non-parametric method. Given a task $\bm{x}^*$, the algorithm finds $K$ samples $\{\bm{x}_k, \bm{y}_k\}_{k=1}^K$ in the database $\{\bm{X},\bm{Y}\}$ where $\{\bm{x}_k\}_{k=1}^K$ are the $K$-nearest to $\bm{x}^*$ according to a chosen metric (in this paper the Euclidean metric is used). It then predicts the corresponding robot path $\bm{y}^*$ by taking the average $\bm{y}^* = \frac{1}{K}\sum_{k=1}^{K} \bm{y}_k$. The method is very simple to implement and it works well if there is a sufficiently dense dataset, but it suffers from the curse of dimensionality; as the dimension of $\bm{x}$ increases, the number of data that needs to be stored increases exponentially. This method is mainly considered as the baseline against the next two methods. 

\subsubsection{Gaussian Process Regressor (GPR)}\label{sec:GPR}

Like $k$-NN, GPR~\cite{rasmusse:book:2006} is a non-parametric method which improves its accuracy as the number of data increases. While having higher computational complexity as compared to $k$-NN, GPR tends to better interpolate, resulting in higher approximation accuracy. Given the database $\{\bm{X}, \bm{Y}\}$, GPR assigns a Gaussian prior to the joint probability of $\bm{Y}$, i.e., $p(\bm{Y}|\bm{X})= \mathcal{N}\big(\bm{\mu}(\bm{X}), \bm{K}(\bm{X},\bm{X})\big)$. $\bm{\mu}(\bm{X})$ is the mean function and $\bm{K}(\bm{X},\bm{X})$ is the covariance matrix constructed with elements $\bm{K}_{ij} =  {k}(\bm{x}_i,\bm{x}_j)$, where ${k}(\bm{x}_i,\bm{x}_j)$ is the kernel function that measures the similarity between the inputs $\bm{x}_i$ and $\bm{x}_j$. In this paper we use Radial Basis Function (RBF) as the kernel function, and the mean function $\bm{\mu}(\bm{X})$ is set to be zero as usually done in GPR. 

To predict the output $\bm{y}^*$ given a new input $\bm{x}^*$, GPR constructs the joint probability distribution of the training data and the prediction, and then conditions on the training data to obtain the predictive distribution of the output, $p({\bm{y}}^* |~\bm{x}^*) \sim \mathcal{N}(\bm{m},\bm{\Sigma})$, where $\bm{m}$ is the posterior mean computed as
\begin{equation}
\bm{m} = \bm{K}({\bm{x}}^*,\bm{X})\, \bm{K}^{-1}(\bm{X},\bm{X})\,  \bm{Y}(\bm{X}), \label{eq:post_mean}
\end{equation}
and $\bm{\Sigma}$ is the posterior covariance which provides a measure of uncertainty on the output. In this work we simply use the posterior mean $\bm{m}$ as the output, i.e., $\bm{y}^* = \bm{m}$.

While having good approximation accuracy, one major limitation with GPR is that it does not scale well with very large datasets. There are variants of GPR that attempt to overcome this problem, e.g., sparse GPR~\cite{quinonero2005unifying} or using Stochastic Variational Inference (SVI)~\cite{hensman2013gaussian}. 
More details on GPR can be found in~\cite{rasmusse:book:2006} and~\cite{bishop2006pattern}.

\subsubsection{Bayesian Gaussian Mixture Regression (BGMR)}\label{sec:DP-GLM}

When using RBF as the covariance function, GPR assumes that the mapping from $\bm{x}$ to $\bm{y}$ is smooth and continuous. When this assumption is met, it performs very well, but otherwise it will yield poor results. For example, when there is discontinuity in the mapping or there are multimodal outputs, GPR tends to average the solutions from both sides of the discontinuity or from both modes. This characteristic is also shared by many other function approximators. To handle discontinuity and multimodality problems, using local models is one of the possible solutions. Each local model can be fit to each side of the discontinuity or to each mode.

Gaussian Mixture Regression (GMR) is an example of such local models approaches~\cite{calinon2016tutorial}. It can be seen as a probabilistic mixture of linear regressions. Given the database $\{\bm{X},\bm{Y}\}$ it can be used to construct the joint probability of $(\bm{x},\bm{y})$ as a mixture of Gaussians
\begin{equation}
p(\bm{x},\bm{y}) = \sum_{k=1}^K \pi_k \, \mathcal{N}(\bm{\mu}_k,\bm{\Sigma}_k),
\end{equation}	
where $\pi_k$, $\bm{\mu}_k$, and $\bm{\Sigma}_k$ are the $k$-th component's mixing coefficient, mean, and covariance, respectively. Given a query $\bm{x}^*$, the conditional probability of the output $\bm{y}^*$ is also a mixture of Gaussians. 

In GMR, the parameters $\pi_k$, $\bm{\mu}_k$ and $\bm{\Sigma}_k$ are determined from the data by Expectation-Maximization method, while the number of Gaussians $K$ is usually determined by the user. Bayesian GMR (BGMR) \cite{pignat2019} is a Bayesian extension of GMR that allows us to estimate the posterior distribution of the mixture parameters (instead of relying on a single point estimate as in GMR). The number of components $K$ can also be automatically determined from the data. As a Bayesian model, BGMR gives priors to the parameters $\pi_k$, $\bm{\mu}_k$ and $\bm{\Sigma}_k$, and computes the posterior distribution of those parameters given the data. In high dimensional problems, the prior reduces the overfitting that commonly occurs with GMR. The prediction $\bm{y}^*$, given the input $\bm{x}^*$, is then computed by marginalizing over the posterior distribution and conditioning on $\bm{x}^*$. The resulting predictive distribution of $\bm{y}$ is a mixture of t-distributions,
\begin{equation}
p(\bm{y}|\bm{x}^*,\bm{X},\bm{Y}) = \sum_{k=1}^K p(k|\bm{x}^*,\bm{X},\bm{Y}) \, p(\bm{y}|k,\bm{x}^*,\bm{X},\bm{Y}),
\label{eq:bgmr_pred}
\end{equation}
where $p(k|\bm{x}^*,\bm{X},\bm{Y})$ is the probability of $\bm{x}^*$ belonging to the $k$-th component of the mixture, and $p(\bm{y}^*|k,\bm{x}^*, \bm{X},\bm{Y})$ is a multivariate t-distribution, the mean of which is linear in $\bm{x}^*$. We can interpret \eqref{eq:bgmr_pred} as $K$ probabilistic linear regression models, each of which has the probability of $p(k|\bm{x}^*,\bm{X},\bm{Y})$. More details about BGMR can be found in~\cite{pignat2019}.

To obtain a point-prediction $\bm{y}^*$ from \eqref{eq:bgmr_pred}, there are several approaches. One of the most used is to take the mean of the predictive distribution in \eqref{eq:bgmr_pred} using moment matching. While this approach can provide smooth estimates (as required in many applications), the same problems as in GPR will appear in the case of discontinuity and multimodality; taking average in those cases will give us poor results. Instead, we propose to take, as the point prediction, the mean\footnote{As in Gaussian distribution, the mean of a multivariate t-distribution is also its mode.} of the component in \eqref{eq:bgmr_pred} having the highest probability, which approximately corresponds to the mode of the multimodal distribution. Alternatively, we can also use the mean of each t-distributions as separate predictions, which gives us several possible solutions. In some cases (e.g., when we would like to retrieve all possible solutions) this approach can be very useful, as will be presented in Section \ref{sec:base_planning}.

\subsubsection{Dimensionality reduction}

In our problem, the path $\bm{y}\in\mathbb{R}^{DT}$ is a vector consisting of the sequence of configurations with dimension $D$ during $T$ time steps, which can be very high. This motivates us to use dimensionality reduction techniques to reduce the dimension of $\bm{y}$. For example, when $T$ is large and the time interval is small, RBF can be used to represent the evolution of each variable as weights of the basis functions. Techniques such as Principal Component Analysis (PCA), Independent Component Analysis, Factor Analysis, and Variational Autoencoder~\cite{bishop2006pattern}~\cite{doersch2016tutorial} can also be used. The mapping to be learned then becomes the mapping from $\bm{x}$ to $\hat{\bm{y}}$, where $\hat{\bm{y}}$ is the projection of $\bm{y}$ to the lower dimensional subspace. The advantage is that the memory required to store the data is reduced significantly, while the approximation performance is maintained or even improved because the important correlations between the variables are preserved. In this work, since the number of time steps is not large, we use PCA to reduce the dimension of $\bm{y}$.

\begin{algorithm}[t]
\small
\caption{Building a Memory of Motion}\label{tab:algo_memmo}
\begin{algorithmic}[0]
\State \textbf{INPUT}: number of samples $N$
\State \textbf{OUTPUT}: the database $\{\bm{X},\bm{Y}\}$ and the function approximator $\bm{f}$
\end{algorithmic}

\begin{algorithmic}[1]
\State $\bm{X}  = [~]$, $\bm{Y} = [~]$
\FOR{$i$ = 1, 2, \dots, $N$}
\State sample a random task $\bm{x}_i$
\State compute the initial guess $\tilde{\bm{y}}_i$ to achieve $\bm{x}_i$ by straight-line motion
\State solve $\bm{x}_i$ using TrajOpt warm-started by $\tilde{\bm{y}}_i$, to obtain the path ${\bm{y}}_i$
	\IF {$\bm{y}_i$ is valid}
	\State add ($\bm{x}_i$, $\bm{y}_i$) to $\{\bm{X},\bm{Y}\}$
	\ENDIF
\ENDFOR

\State apply PCA to $\bm{Y}$ to obtain $\hat{\bm{Y}}$ (\textit{Optional})
\State train the function approximator $\bm{f}$ on $\{\bm{X},\bm{Y}\}$ \quad (\textit{or on} $\{\bm{X},\hat{\bm{Y}}\}$ \textit{if PCA is used})

\end{algorithmic}
\end{algorithm}

\begin{algorithm}[t]
\small
\caption{Using the Memory as a Metric}\label{tab:metric}
\begin{algorithmic}[0]
\State \textbf{INPUT}: A list of goals $\{\bm{x}_j\}_{j=1}^M$, a function approximator $\bm{f}$
\State \textbf{OUTPUT}: The optimal goal $\bm{x}^*$ and the corresponding path $\bm{y}^*$
\end{algorithmic}

\begin{algorithmic}[1]
\FOR{$j$ = 1, 2, \dots, $M$}
	\State compute the initial guess $\tilde{\bm{y}}_j$ = $\bm{f}(\bm{x}_j)$
	\State compute the cost $\ell(\tilde{\bm{y}}_j)$
\ENDFOR
\State $j^* \gets$ $\argmin_{j}$ $\ell(\tilde{\bm{y}}_j)$
\State $\bm{x}^*$ $\gets$ $\bm{x}_{j^*}$ 
\State $\tilde{\bm{y}}^*$ $\gets$ $\tilde{\bm{y}}_{j^*}$ 
\State solve $\bm{x}^*$ using TrajOpt warm-started by $\tilde{\bm{y}}^*$, to obtain the path ${\bm{y}}^*$
\end{algorithmic}
\end{algorithm}

\begin{algorithm}[t]
\small
\caption{Ensemble Method}\label{tab:ensemble}
\begin{algorithmic}[0]
\State \textbf{INPUT}: Task $\bm{x}^*$, a list of function approximators $\{\bm{f}_j\}_{j=1}^M$
\State \textbf{OUTPUT}: The path $\bm{y}^*$ that accomplishes the task $\bm{x}^*$
\end{algorithmic}

\begin{algorithmic}[1]
\FORALLP{$j$ = 1, 2, \dots, $M$}
	\State compute the initial guess $\tilde{\bm{y}}_j$ = $\bm{f}_j(\bm{x}^*)$
	\State solve $\bm{x}^*$ using TrajOpt warm-started by $\tilde{\bm{y}}_j$, to obtain the path ${\bm{y}}_j$
	\IF {$\bm{y}_j$ is valid}
	\State $\bm{y}^*$ = $\bm{y}_j$  	
	\State Terminate the parallel execution
	\ENDIF
\ENDFAP

\end{algorithmic}
\end{algorithm}

\subsection{Using the Memory as a Metric}
\label{sec:metric}

In some planning problems, there can be several alternative goals to be achieved. For example, in robot drilling task~\cite{suarez2018robotsp}, the orientation around the drilling axis is free (the number of possible goals is infinite). A naive way is to choose one of the goals randomly, plan the motion, and if it fails then select another goal. While this is simple to implement, it does not make use of the benefit of having multiple goals. Another method is to plan the paths to each goal and select the one having the smallest cost, but this is computationally expensive. It will be useful, therefore, to have a metric that measures the cost to a given goal. Our idea is to use the memory of motion as the metric.

In Section \ref{sec:building_memmo}, function approximators were trained to predict an initial guess to achieve a task $\bm{x}$. The possible goals can then be formulated as multiple tasks $\{\bm{x}_0, \bm{x}_1, \dots, \bm{x}_{M-1}\}$. For each task $\bm{x}_i$, the function approximator predicts the initial guess $\tilde{\bm{y}}_i$ corresponding to the task, and the cost $\ell(\tilde{\bm{y}}_i)$ can be computed. The initial guess $\tilde{\bm{y}}_i^*$ and the corresponding task $\bm{x}_i^*$ with the lowest cost is then taken as the chosen goal to be given to the trajectory optimizer. Since the cost computation (the total discrete velocity in \eqref{eq:discrete_vel}) can be done quickly relative to optimization time, this approach can yield significant improvements to the trajectory optimizer performance.

\subsection{Using Ensemble Method to Provide Warm-Start}
\label{sec:ensemble}

In machine learning, methods such as AdaBoost~\cite{freund1997decision} and Random Forests~\cite{breiman2001random} have shown that using an ensemble of methods often yields improved performances as compared to choosing a single method. We propose to use an ensemble method where we run multiple trajectory optimizations in parallel, each one warm-started by one of the function approximators in Section \ref{sec:building_memmo}, and once one of them finds a successful path the others are terminated. Since each function approximator has different learning characteristics, combining them in this way can significantly improve the motion planning performance. The method in Section \ref{sec:metric} can also be used as one of the ensemble's component.

\section{Experiments}
\label{sec:experiments}

To evaluate the proposed method, we consider several examples of motion planning for PR2 and Atlas robots. TrajOpt~\cite{schulman2013finding} is used as the trajectory optimizer to be warm-started. 
The output is the robot path that accomplishes the given task. In this paper we only work with robot path as the output, but the method can also be applied to robot trajectory. 

We consider 5 motion planning cases presented in ascending order of complexity. 
Each case is chosen to demonstrate certain characteristics of the proposed method. 
For each case, we follow the following procedures. 
First we generate the dataset by randomly sampling $N_\text{train}$ tasks from a uniform distribution and run TrajOpt to find the paths achieving the tasks. 
The number of time steps $T$ is set to $30$, except for Atlas ($T=15$). 
In all cases, the cost is defined as the discrete velocity of the states, as defined in \eqref{eq:discrete_vel}. The number of $N_\text{train}$ is different for each case, depending on the complexity of the task. The function approximators are then trained with or without PCA using the dataset. We heuristically set $50$ components for the PCA; for the $k$-NN, we use $K=1$. 

To validate the performance, we sample $N_\text{test}$ random tasks and use the various methods to warm-start TrajOpt. The solutions are compared in terms of \emph{convergence time}, \emph{success rate} and \emph{cost}. The planning is considered successful if the solution is feasible. The comparison results are presented in the Tables \ref{tab:base_left}-\ref{tab:atlas_result}. The values are averaged over $N_\text{test}$ tasks, and the standard deviation is also given for the convergence time and the cost. In the presented results, we use the label `STD' to refer to the solution obtained by warm-starting the solver with a straight-line path (via waypoint, if any), and the names of the function approximators for the rest. The subscript `PCA' is added when PCA is used. The query time for predicting the warm-starts by each method is negligible w.r.t. the convergence time, i.e. less than 5~ms for most methods, except for BGMR without PCA (around 20ms), so they are not included in the comparison. The codes to run the experiments are provided in \url{https://github.com/teguhSL/memmo_for_trajopt_codes}, and the videos are submitted as supplementary file. 

\renewcommand{\arraystretch}{0.8}
\begin{table}[!t]
	\centering
      \caption{Base motion planning, one waypoint.}
      
	\begin{tabular}{ l  c  c   c }
		\toprule
		\multirow{2}{1cm}{\bf{Method}} & $\textbf{Success}$ & $\textbf{Conv.}$ & $\textbf{Cost}$ \\  
			 				  &$\textbf{(\%)}$  & $\textbf{time (s)}$ & $\textbf{(rad/s)}$ \\  
       \midrule
       STD     	& 80.0 	& 0.55$\pm$0.29 	& 1.37$\pm$0.37 \\
	   $k$-NN      	& 93.0 	& 0.35$\pm$0.20 	& 1.45$\pm$0.47 \\
	   GPR     	& 96.0 	& 0.37$\pm$0.15 	& \textbf{1.32$\pm$0.36} \\
	   BGMR     & \textbf{97.0} 	& \textbf{0.32$\pm$0.14} 	& 1.34$\pm$0.35 \\
	
		\bottomrule
	\end{tabular}
	\label{tab:base_left}
\end{table}

\renewcommand{\arraystretch}{0.8}
\begin{table}[!t]
	\centering
     \caption{Base motion planning, two waypoint.}
      
	\begin{tabular}{ l c c c}
		\toprule
		\multirow{2}{1cm}{\bf{Method}} & $\textbf{Success}$ & $\textbf{Conv.}$ & $\textbf{Cost}$ \\  
			 				  &$\textbf{(\%)}$  & $\textbf{time (s)}$ & $\textbf{(rad/s)}$ \\  
		\midrule
		STD     	& 79.0 	& 0.53$\pm$0.23 	& 1.43$\pm$0.37 \\
		$k$-NN      & \textbf{95.0} 	& 0.32$\pm$0.16 	& 1.53$\pm$0.62 \\
		GPR     	& 00.0 	& - 	& - 	 \\
		BGMR~~~~~~  & 94.0 	& \textbf{0.31$\pm$0.15} 	& \textbf{1.33$\pm$0.40} \\
		\bottomrule
	\end{tabular}
	\label{tab:base_both}
\end{table}

\begin{figure}[t!]
\centering
\subfloat[][]{\includegraphics[width=0.245\columnwidth]{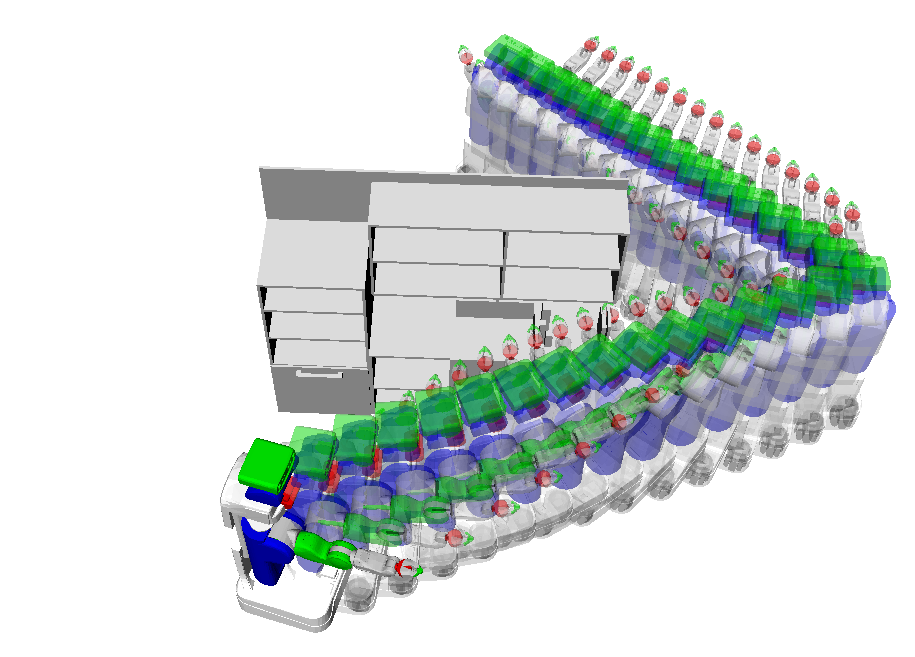}\label{fig:straight_line}} 
\hfill
	\subfloat[][]{\includegraphics[width=0.245\columnwidth]{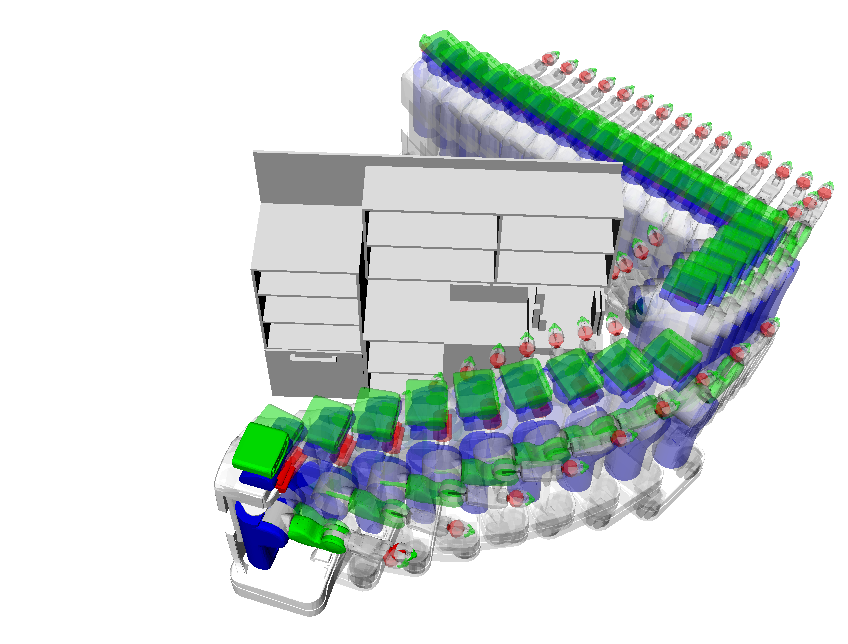}\label{fig:nn}}
\hfill
\subfloat[][]{\includegraphics[width=0.245\columnwidth]{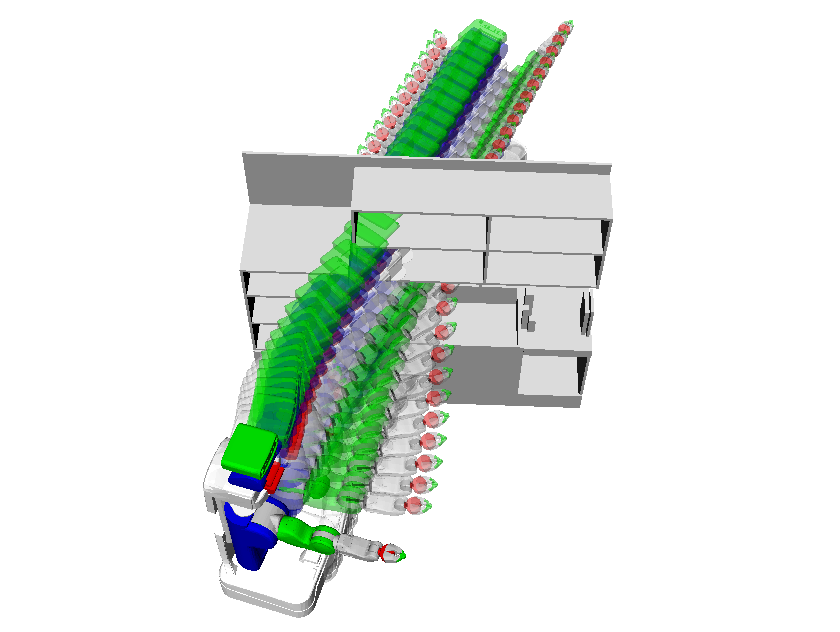}\label{fig:gpr}}
\hfill
\subfloat[][]{\includegraphics[width=0.245\columnwidth]{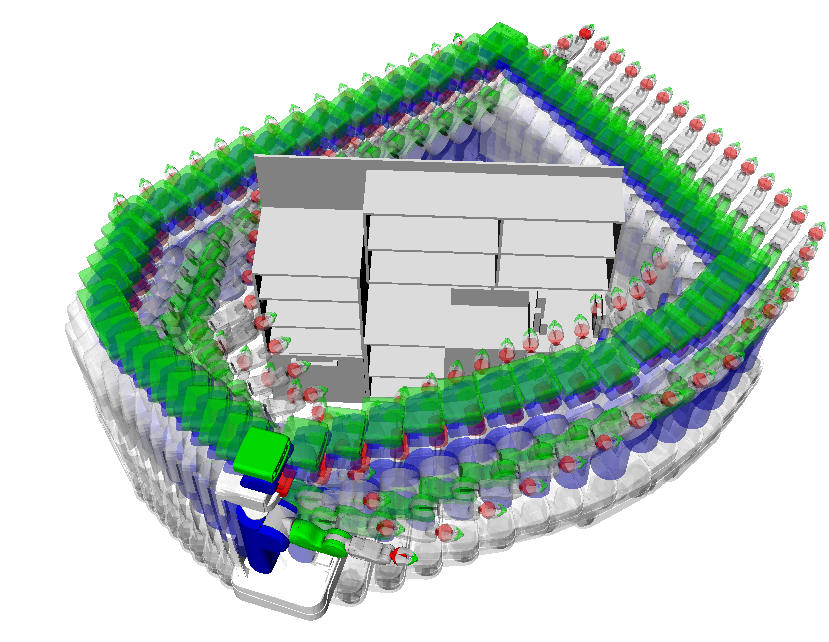}\label{fig:dp_gmm}}
\caption{
Motion planning for the PR2 mobile base. Warm-start produced by \protect\subref{fig:straight_line}~straight-line with waypoint, \protect\subref{fig:nn}~$k$-NN, \protect\subref{fig:gpr}~GPR and \protect\subref{fig:dp_gmm}~BGMR. }
\label{fig:base_planning}
\end{figure}

\subsection{Base motion planning}
\label{sec:base_planning}

The task is to plan the motion for the PR2 mobile base from a random pose in front of the kitchen to another random pose behind the kitchen (Fig.~\ref{fig:pr2_example_kitchen}). 
In this case, the state $\bm{q}$ is the 3 DoF planar pose of the base. The task descriptor is then $\bm{x} = (\bm{q}_\text{init}^\top, \bm{q}_\text{goal}^\top)^\top$. The database is constructed with $N_\text{train} = 200$ samples and the evaluation is performed with $N_\text{test} = 100$. Although this is an easy problem, TrajOpt actually finds it difficult to solve without a proper initialization. 
For example, initializing TrajOpt with a straight-line interpolation from $\bm{q}_\text{init}$ to $\bm{q}_\text{goal}$ never manages to find a feasible solution because it results in a path that moves the robot through the kitchen while colliding, and the solver get stuck in poor local optima due to the conflicting gradients. To obtain better initialization for building the database, we initialize TrajOpt with two manually chosen waypoints on the left and on the right of the kitchen ($\bm{q}_\text{left}$ and $\bm{q}_\text{right}$, respectively).

We consider two cases of building the database: in the first one, we only use $\bm{q}_\text{right}$ as waypoint, while in the second we use both $\bm{q}_\text{left}$ and $\bm{q}_\text{right}$. We initialize TrajOpt with the straight-line motion from $\bm{q}_\text{init}$ to the waypoint and from the waypoint to $\bm{q}_\text{goal}$. With this setting we build the database, train the function approximators, and obtain the results as shown in Table~\ref{tab:base_left} and~\ref{tab:base_both}.  

In the first case, the mapping from $\bm{x}$ to $\bm{y}$ is unimodal because all movements go through the right. Table \ref{tab:base_left} shows that the performance of $k$-NN, GPR and BGMR are quite similar. In the second case, however, the output is multimodal because the database contains two possible ways (modes) to accomplish the same task. This affects GPR significantly (see Table \ref{tab:base_both}), as GPR averages both modes and outputs a path that goes through the kitchen, while $k$-NN and BGMR are not affected. $k$-NN does not average the modes because we use $K=1$, while BGMR overcomes the multimodality by constructing local models for each mode automatically.
 
Fig.~\ref{fig:base_planning} shows the examples of warm-starts produced by each method in the second case. As expected, GPR provides a warm-start that goes through the kitchen (hence 0 success rate). With BGMR, if we retrieve the components with the two highest probability, both possible solutions are obtained.

\renewcommand{\arraystretch}{0.8}
\begin{table}[!t]
     \caption{Planning from fixed $\bm{q}_\text{init}$ to random $\bm{q}_\text{goal}$.}      
      \centering
	\begin{tabular}{ l c c  c }
		\toprule
		\multirow{2}{1cm}{\bf{Method}} & $\textbf{Success}$ & $\textbf{Conv.}$ & $\textbf{Cost}$ \\  
			 				  &$\textbf{(\%)}$  & $\textbf{time (s)}$ & $\textbf{(rad/s)}$ \\  
	    \midrule
STD     	& 80.0 	& 0.77$\pm$0.37 	& 1.83$\pm$0.61 	 \\
$k$-NN      	& 91.2 	& \textbf{0.58$\pm$0.29 } 	& 1.93$\pm$0.69 	 \\
GPR     	& 92.4 	& 0.65$\pm$0.25	& 1.84$\pm$0.57  	 \\
GPR$_\text{PCA}$ 	& \textbf{92.8} 	& 0.66$\pm$0.26  	& \textbf{1.83$\pm$0.57} 	 \\
BGMR 	& 88.8 	& 0.64$\pm$0.26 	& 1.85$\pm$0.56 	 \\
BGMR$_\text{PCA}$ 	& 92.0 	& 0.67$\pm$0.26 	& 1.84$\pm$0.58 	 \\
		\bottomrule
	\end{tabular}
	\label{tab:fix_init}
\end{table}

\begin{table}[!t]   
      \centering
    \caption{Planning from random $\bm{q}_\text{init}$ to $\bm{q}_\text{goal}$.}
    \centering
	\begin{tabular}{ l c c  c}
			\toprule
			\multirow{2}{1cm}{\bf{Method}} & \bf{Success} & \bf{Conv.} & \bf{Cost} \\  
			 				               & \bf{(\%)}    & \bf{time (s)} & \bf{(rad/s)} \\
			\midrule
			STD     	& 75.2 	& 0.82$\pm$0.43 	& \textbf{1.31$\pm$0.74} \\
			$k$-NN      & 65.6 	& 1.16$\pm$0.58 	& 1.55$\pm$0.88 \\
			GPR     	& 85.6 	& 0.85$\pm$0.39 	& 1.32$\pm$0.74 \\
			GPR$_\text{PCA}$ 	& 88.0 	& \textbf{0.81$\pm$0.36} 	& 1.33$\pm$0.73 \\
			BGMR 	& 84.0 	& 0.81$\pm$0.40 	& 1.34$\pm$0.76 \\
			BGMR$_\text{PCA}$ 	& 78.3 	& 0.88$\pm$0.42 	& 1.39$\pm$0.78 \\
			Waypoints 	& 94.0 	& 1.52$\pm$0.67 	& 1.83$\pm$1.34 \\
			Ensemble 	& \textbf{97.2} 	& 1.06$\pm$0.41 	& 1.42$\pm$0.82 \\
		\bottomrule
	\end{tabular}
	\label{tab:random_init}
\end{table}

\subsection{Planning from a fixed initial configuration to a random goal configuration}
\label{sec:fix_init}

Here $\bm{q}$ consists of $14$ joint angles of the two $7$ DoFs arms of PR2. The task $\bm{x}$ is to move from a fixed $\bm{q}_\text{init}$ to a random goal configuration $\bm{q}_\text{goal}$ (i.e. $\bm{x} = \bm{q}_\text{goal}$). The database is constructed with $N_\text{train} = 500$, and the evaluation results with $N_\text{test} = 250$ are presented in Table~\ref{tab:fix_init}. 

Since each PR2 arm is redundant, the path from $\bm{q}_\text{init}$ to $\bm{q}_\text{goal}$ can be multimodal, which may pose a problem for GPR. 
However, Table \ref{tab:fix_init} shows that GPR and BGMR perform similarly. 
This is due to the fact that although redundant robots can achieve a goal configuration in many different ways, planning using optimization here results in similar motions for similar goal configurations. The use of PCA does not improve the performance significantly, but it still helps to reduce the size of the data. In this case, for each path it reduces the number of variables from $30\times 14$ ($D\times T$) to $50$ (number of PCA components), more than 8 times reduction while maintaining the performance.

\subsection{Planning from a random initial configuration to a random goal configuration}
\label{sec:random_init}

To proceed with a more complex case, the task here is to plan a path from a random initial configuration $\bm{q}_\text{init}$ to a random target $\bm{q}_\text{goal}$.
The task $\bm{x}$ consists of the initial and goal configurations, $\bm{x} = {(\bm{q}_\text{init}^\top , \bm{q}_\text{goal}^\top)}^\top$. 
The database is constructed with $N_\text{train}=500$ and evaluated with $N_\text{test}=250$. 
The result is presented in Table \ref{tab:random_init}. 

$k$-NN performs poorly here, similar to STD, due to the dimension of the input space $\bm{x}$ that is much larger as compared to Section \ref{sec:fix_init}.
To achieve good performance, $k$-NN requires a much denser dataset. GPR outperforms BGMR by a wide margin.

The last row of Table~\ref{tab:random_init} shows the result of the ensemble method described in Section \ref{sec:ensemble}. Given an input $\bm{x}^*$, the method uses all function approximators to provide different warm-starts, each of which is used to initialize an instance of TrajOpt in parallel. Once a valid solution is obtained, the other instances of TrajOpt are terminated. This method results in a huge boost of the success rate, with comparable convergence time and cost to the other methods. As comparison, we also include here the standard multiple initializations suggested by TrajOpt (labeled as `waypoints`). Each initialization is created by interpolating through a waypoint that is manually defined. While the success rate is high, the convergence time and the cost increase significantly. On the contrary, each initialization in the ensemble method has a good probability of being close to the optimal solution, resulting in lower cost and convergence time.

\renewcommand{\arraystretch}{0.8}
\begin{table}[t]
\centering
\caption{Planning from fixed $\bm{q}_\text{init}$ to random Cartesian goal.}
\begin{tabular}{l c c c}
\toprule
$\textbf{Method}$ & $\textbf{Success (\%)}$ & $\textbf{Conv. time (s)}$ & $\textbf{Cost (rad/s)}$ \\  \midrule
STD     	& 65.2 	& 1.10$\pm$0.62 	& 1.86$\pm$0.86 \\
$k$-NN      	& 73.6 	& 1.28$\pm$0.96 	& 1.84$\pm$0.81 \\
GPR     	& 66.4 	& 1.81$\pm$0.96 	& 1.87$\pm$0.87 \\
GPR$_\text{PCA}$ 	& 66.8 	& 1.68$\pm$0.98 	& 1.78$\pm$0.83 \\
BGMR 	& 74.4 	& 1.37$\pm$0.82 	& 1.82$\pm$0.86 \\
BGMR$_\text{PCA}$ 	& 77.2 	& 1.33$\pm$0.75 	& 1.84$\pm$0.80 \\
METRIC GPR$_\text{PCA}$ 	& 86.8 	& \textbf{0.70$\pm$0.30} 	& \textbf{1.49$\pm$0.56} \\
Ensemble 	& \textbf{98.0} 	& 1.50$\pm$0.60 	& 1.60$\pm$0.68 \\
\bottomrule
\end{tabular}
\label{tab:cartesian}
\end{table}

\renewcommand{\arraystretch}{0.8}
\begin{table}[t]
\centering
\caption{Planning the motion of Atlas from fixed $\bm{q}_\text{init}$ to random Cartesian goal.}
\begin{tabular}{l c c c}
\toprule
$\textbf{Method}$ & $\textbf{Success (\%)}$ & $\textbf{Conv. time (s)}$ & $\textbf{Cost (rad/s)}$ \\  \midrule
STD     	& 50.8 	& 6.31$\pm$3.90 	& 0.12$\pm$0.07 \\
$k$-NN      & 58.8 	& 1.48$\pm$1.39 	& 0.11$\pm$0.06 \\
GPR     	& 54.4 	& \textbf{1.29$\pm$1.09} 	& \textbf{0.10$\pm$0.05} \\
GPR$_\text{PCA}$    	& 60.0 	& 1.54$\pm$1.46 	& 0.11$\pm$0.05 \\
BGMR    	& 56.4 	& 1.32$\pm$1.57 	& \textbf{0.10$\pm$0.05} \\
BGMR$_\text{PCA}$   	& 58.0 	& 1.36$\pm$1.16 	& 0.11$\pm$0.06 \\
Ensemble 	& \textbf{71.2} 	& 1.46$\pm$1.40 	& 0.12$\pm$0.06 \\
\bottomrule
\end{tabular}
\label{tab:atlas_result}
\end{table}

\subsection{Planning to Cartesian goals from a fixed initial configuration}
\label{sec:cartesian}

In Section \ref{sec:fix_init} and \ref{sec:random_init}, we use TrajOpt to plan to goals in configuration space. In practical situations, however, the task is often to reach a certain Cartesian pose using the end-effector (e.g., to pick an object on the shelf), instead of planning to a specific joint configuration. One way to solve this problem is to first compute a configuration that achieves the Cartesian pose using an inverse kinematic solver and plan to this configuration, but it does not make use of the flexibility inherent in the task. TrajOpt has an option to plan directly to a Cartesian goal, but it typically requires longer convergence time and lower success rate than planning to a joint configuration goal. 

We present two approaches to use the memory of motion in this problem. In the first approach, we rely on the similar procedure as in previous cases: we formulate the task as $\bm{x} = {({\bm{p}}^\top_\text{left}, {\bm{p}}^\top_\text{right})}^\top$ where $\bm{p}_\text{left}$ and $\bm{p}_\text{right}$ are the Cartesian positions of the right and left hand of PR2. The database is then constructed with $N_\text{train}=1000$ and the function approximators are trained. In this approach, TrajOpt plans to a Cartesian goal directly. The second approach relies on the fact that a Cartesian goal corresponds to multiple goals in configuration space. In Section~\ref{sec:fix_init} we have already constructed several function approximators that can predict an initial guess $\tilde{\bm{y}} = \tilde{\bm{q}}_{0:T}$, given a goal $\bm{q}_\text{goal}$ in configuration space. The second approach uses one of them as a \textit{metric} (Sect.~\ref{sec:metric}) to choose between the different goals in configuration space. First, given a Cartesian goal $\bm{x}$, we run an inverse kinematic solver to find $M = 5$ joint configurations that satisfy this pose. For each joint configuration, we use the function approximator to predict the initial guess of the robot path to reach that configuration, and we compute the cost of that path. Finally, the goal configuration and the path with the lowest cost are chosen, and TrajOpt is run to reach this goal configuration with the given path as the warm-start. Note that in this second approach, TrajOpt plans to a joint configuration instead of a Cartesian goal. For this approach we choose the method $\text{GPR}_\text{PCA}$ from the Section \ref{sec:fix_init}, and use the term `$\text{METRIC GPR}_\text{PCA}$' to differentiate from the first approach (denoted in standard notation).

We present the results in Table \ref{tab:cartesian} with $N_\text{test}=250$. Among the methods using the first approach, we note that BGMR yields better result than GPR because the mapping from the Cartesian goal $\bm{x}$ to the robot path $\bm{y}$ here is multimodal, as planning to a Cartesian pose has more redundancy as compared to planning to a joint configuration. This again demonstrates that BGMR handles multimodal output better than GPR. However, the second approach $\text{METRIC GPR}_\text{PCA}$ outperforms even BGMR. The improvement over the first approach is very significant in all three criteria. This demonstrates that using the memory as a metric to choose the optimal goal results in large improvements. We point out that the additional computational time required to find $M=5$ IK solutions and the corresponding warm-starts is only around $0.1$~s, which is negligible compared to the convergence time. Finally, we use the ensemble method that uses all function approximators in parallel, including $\text{METRIC GPR}_\text{PCA}$. This boosts the success rate to 98\%.

\subsection{Planning whole-body motion for an Atlas robot}
\label{sec:atlas}

Finally, we also applied our method for planning the motion of the 34-DoFs Atlas robot (28-DoFs joints and 6-DoFs root pose). We consider the same task as in Section \ref{sec:cartesian}, i.e. planning from a fixed initial configuration to a random Cartesian pose, in this case chosen to be the location of Atlas' right hand. 
The task $\bm{x} = (p_x, p_y, p_z)^\top$ corresponds to the target position of Atlas' right hand, while the orientation is not constrained. The feet location are fixed, while the Zero Moment Point (ZMP) is constrained to be between the two feet location. We use here the first approach as explained in Section \ref{sec:cartesian}, i.e. treating it as a regression problem where the input $\bm{x}$ is the Cartesian goal and the output $\bm{y}$ is the trajectory, and use the various function approximators to predict the initial guesses. The database is constructed with $N_\text{train} = 1000$ and the evaluation is performed with $N_\text{test} = 250$. The results are presented in Table \ref{tab:atlas_result}.

$k$-NN performs quite well, as the input size of $\bm{x}$ is small (the position of the hand is constrained to be inside the shelf). Unlike in Section \ref{sec:cartesian}, the performance of GPR and BGMR are quite similar, although the goals are also in the Cartesian space. This is due to the difference in the implementation; in Section \ref{sec:cartesian}, given a Cartesian goal, we use an inverse kinematic solver to calculate the joint configuration that satisfies this goal, and calculate the initial guess as straight-line interpolation from the fixed initial configuration to the goal configuration. This initial guess is used when building the database. Due to the redundancy of the PR2 dual arm, similar Cartesian goals can correspond to very different joint configurations, resulting in the multimodality of the solutions in the database. In this Atlas experiment, however, we do not provide initial guesses to TrajOpt when building the database, so TrajOpt always tries to solve the problem with zero initialization. This results in more uniform solutions, and hence GPR can still perform quite well. Finally, using the ensembe method again shows superior results, giving us an increase of the success rate by more than 10\%.

Planning for such high DoFs problem with many constraints (feet location, ZMP constraint, kinematic constraint) requires quite a lot of computational time ($\sim$ 6.3 s in average without warm-start). Using the memory of motion in this complex task further exemplify the benefit of the approach, as our method speeds up the computational time significantly by more than four times faster. We note that the tasks are sampled randomly, and there is no guarantee that the task is indeed feasible. This explains why even the best method (i.e. the ensemble method) only achieves $\sim$ 70\% success rates.    

\section{Discussions}
\label{sec:discussion}

\begin{figure*}
\centering
\subfloat[][]{\includegraphics[height=0.45\columnwidth]{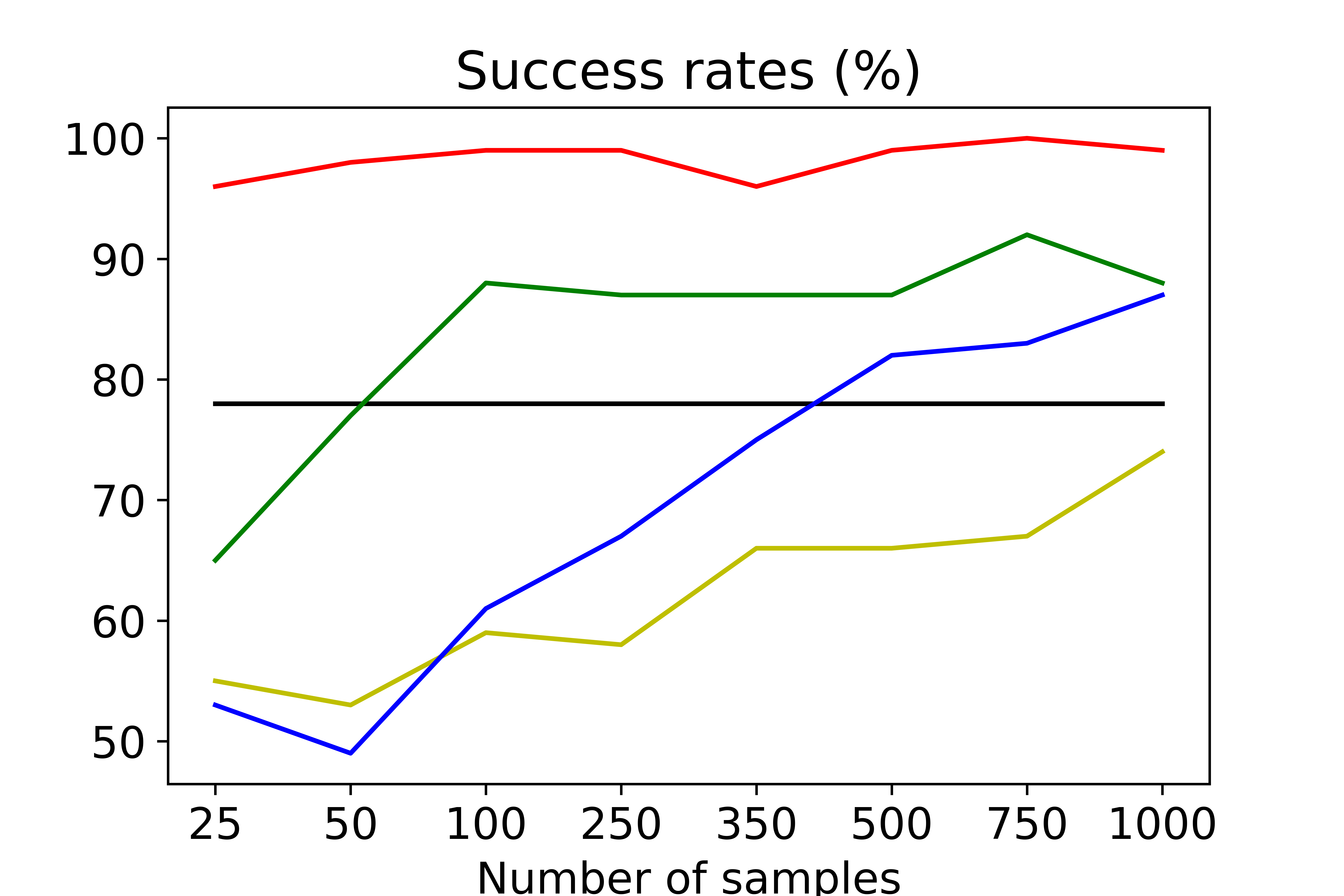}\label{fig:success_rates}}
\hfill
\subfloat[][]{\includegraphics[height=0.45\columnwidth]{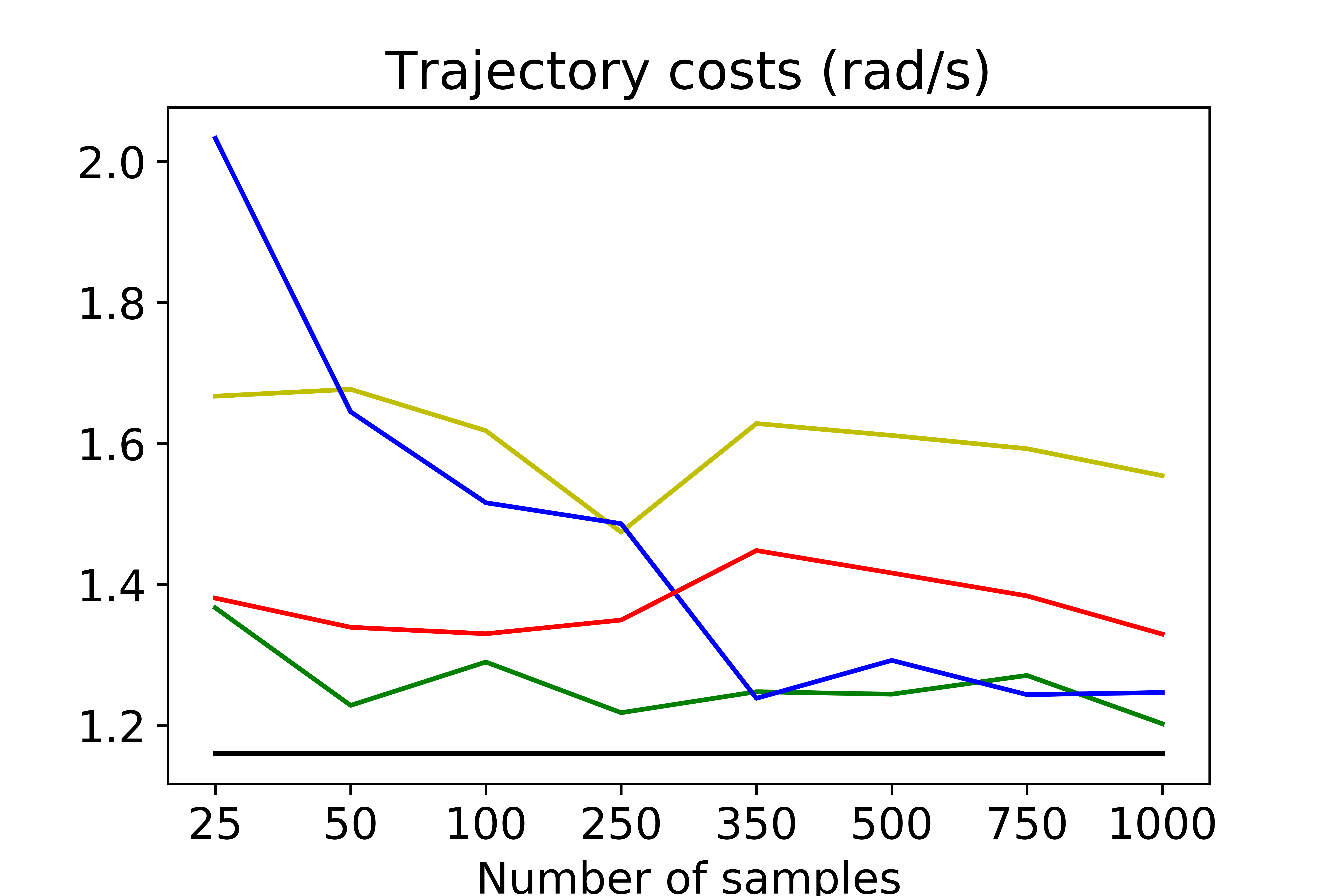}\label{fig:traj_costs}}
\hfill
\subfloat[][]{\includegraphics[height=0.45\columnwidth]{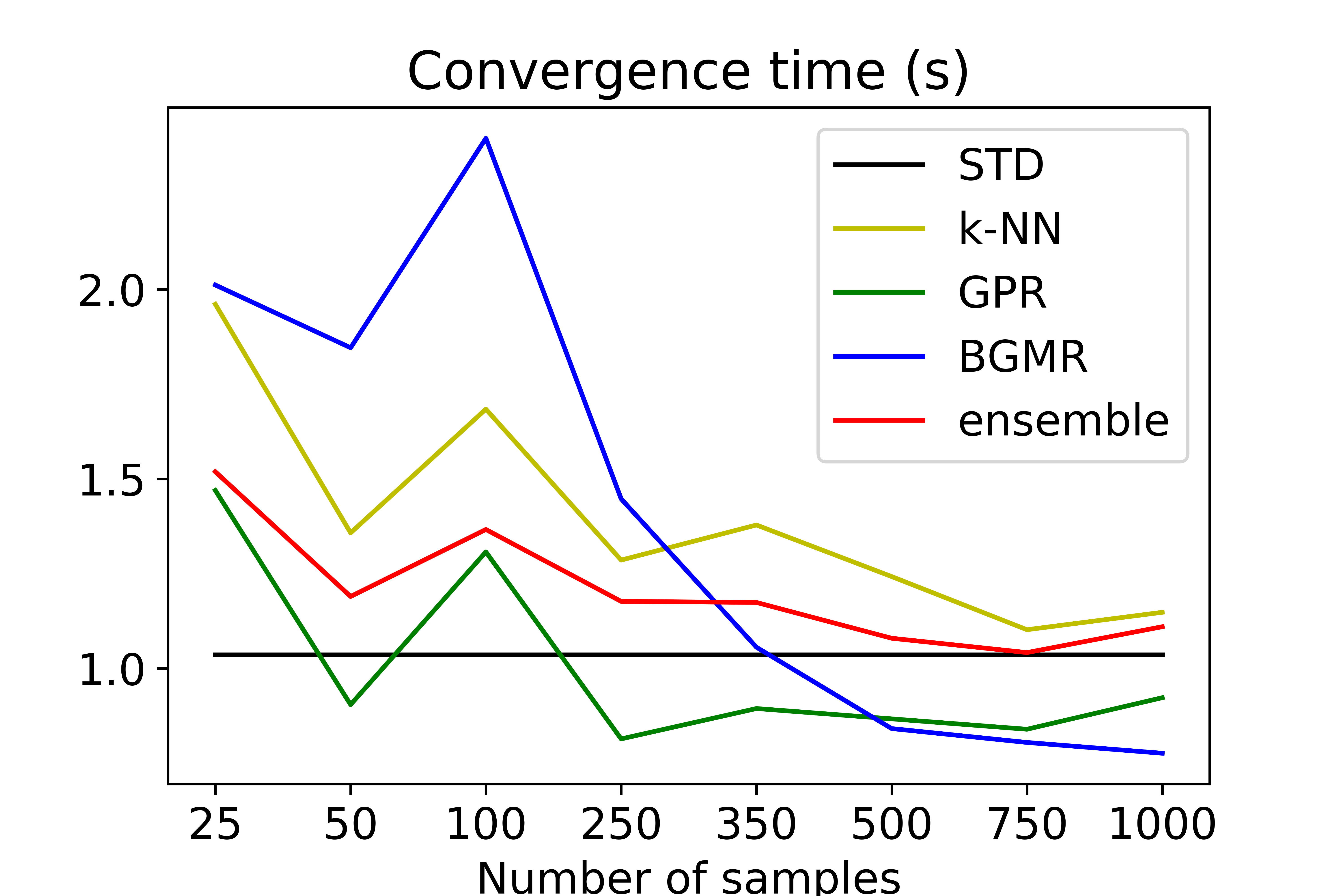}\label{fig:conv_time}}
\caption{Performance comparison against the number of training samples $N_\text{train}$}
\label{fig:data_comparison}
\end{figure*}

\subsection{Choice of function approximators}
In Section \ref{sec:experiments}, we have compared the performance of $k$-NN, GPR and BGMR over different tasks, and shown that they have different characteristics. When the dataset is quite dense or the input space is small, $k$-NN usually manages to obtain good performance (as shown in Section \ref{sec:base_planning}, \ref{sec:fix_init}, and \ref{sec:cartesian}), while for larger input space (Section \ref{sec:cartesian}) it does not yield good results. GPR performs the best when the output is unimodal (Section \ref{sec:fix_init} and \ref{sec:random_init}), while for multimodal output BGMR has a better performance than GPR (Section \ref{sec:cartesian}). This comparison can guide us to select the best method for each task. However, it may not be obvious whether a given task (and its solution) is unimodal or multimodal (e.g. compare Section \ref{sec:cartesian} and \ref{sec:atlas}). A better way is to combine the different methods via an ensemble method, as we have shown in this paper.

\subsection{Data requirement}
In Fig.~\ref{fig:data_comparison}, we plot the performance of various methods against the number of training samples, with STD given as the baseline. We choose the task in Section \ref{sec:random_init}, since it has the largest input space among the other tasks. It is interesting that when the training size is small, GPR performs quite well, while $k$-NN and BGMR are even worse than STD. As training size increases, $k$-NN and BGMR start to approach the performance of GPR. On the contrary, the performance of the ensemble method is quite stable even when the training size is small. As the training size grows, its convergence time decreases, while the success rate is already high even when the training size is small.

\subsection{Ensemble method}
Using an ensemble method for motion planning has been explored in \cite{choudhury2015planner}, which uses an ensemble of motion planners. While such approach also manages to boost the performance successfully, it is not easy to design and set up several motion planners for a given task. On the contrary, many function approximators are available and can be used easily, since our problem is formulated as a standard regression problem. We only need to configure one motion planner (in this work, TrajOpt, but other optimization frameworks can also be used) for a given task, unlike in \cite{choudhury2015planner}. Another benefit of our ensemble method is that each of the ensemble's component starts from an initial guess that has good probability of being close to the optimal solution. This reduces the average computational time, as we have shown by comparing it against the multiple waypoints initialization in Table \ref{tab:random_init}. 

\subsection{Dynamic environment}
In this work we assume that the environment is static, so that the trajectories previously planned remain valid. When the environment changes, a new memory of motion has to be built. For the simple example in Section \ref{sec:base_planning}, building the memory takes only $\sim$3 minutes of computational time, but complex example such as Section \ref{sec:atlas} takes $\sim$3 hours. While paralellization can be used to speed up the building process, more effective strategies would be interesting to explore. In \cite{Yiming17}, an efficient way of updating a dynamic roadmap when the environment changes is presented. Such method can possibly be used to modify the existing memory of motion, so that we do not have re-build from scratch but only modify those affected. Alternatively, when the environment largely remain the same but a few obstacles are moving (as in many real tasks), we can include these obstacles' locations as additional inputs to the regression problem, at the expense of larger input size. We will explore these ideas in our future work.

\section{Conclusion}
\label{sec:conclusion}

We have presented an approach to build a memory of motion to warm-start trajectory optimization solver, and demonstrate through experiments with PR2 and Atlas robots that the warm-start can improve the solver's performance. Function approximators and dimensionality reduction are used to learn the mapping between the task descriptor and the corresponding robot path. Three function approximators are considered: $k$-NN as baseline, GPR, and BGMR, and their different characteristics have been discussed. The use of PCA also improves the solution, although not very significantly, while reducing the memory storage.  We have also shown that we can use the memory of motion as a metric to choose optimally between several alternative goals, and this results in a significantly improved performance for the case of Cartesian goal planning. Finally, the different function approximators can be combined as an ensemble method, which boosts the success rate significantly.


\IEEEtriggeratref{10}

\bibliographystyle{IEEEtran}
\bibliography{main}

\end{document}